\documentclass[11pt]{article}
\usepackage{acl2015}
\usepackage{times}
\usepackage{url}
\usepackage{latexsym}
\usepackage{times}
\usepackage{url}
\usepackage{graphicx}
\usepackage{epstopdf}
\usepackage{amsfonts}
\usepackage{amssymb}
\usepackage{amsmath}
\usepackage{verbatim}
\usepackage{multirow}
\usepackage{subfigure}
\usepackage{array}
\usepackage{fp}
\usepackage{makecell}
\usepackage{flushend}
\usepackage{cases}
\usepackage{color}
\usepackage{bbm}
\usepackage{mathrsfs}
\usepackage{amsbsy}

\date{}
\title{Hierarchical Recurrent Attention Network for Response Generation}
\author{
	Chen Xing$^{12}$\thanks{The work was done when the first author was an intern in Microsoft Research Asia.}~~, Wei Wu$^3$~~, Yu Wu$^4$~~, Ming Zhou$^3$~~, Yalou Huang$^{12}$~~, Wei-Ying Ma$^3$\\
	$^1$College of Computer and Control Engineering, Nankai University, Tianjin, China\\
	$^2$College of Software, Nankai University, Tianjin, China\\
	$^3$~~~~Microsoft Research, Beijing, China\\
	$^4$State Key Lab of Software Development Environment, Beihang University, Beijing, China\\
	\{v-chxing,wuwei,v-wuyu,mingzhou,wyma\}@microsoft.com ylhuang@nankai.edu.cn
}
\begin{document}
	\maketitle
	\begin{abstract}
		We study multi-turn response generation in chatbots where a response is generated according to a conversation context.  
	Existing work has modeled the hierarchy of the context, but does not pay enough attention to the fact that words and utterances in the context are differentially important. As a result, they may lose important information in context and generate irrelevant responses. We propose a hierarchical recurrent attention network (HRAN) to model both aspects in a unified framework. In HRAN, a hierarchical attention mechanism attends to important parts within and among utterances with word level attention and utterance level attention respectively. With the word level attention, hidden vectors of a word level encoder are synthesized as utterance vectors and fed to an utterance level encoder to construct hidden representations of the context. The hidden vectors of the context are then processed by the utterance level attention and formed as context vectors for decoding the response.  Empirical studies on both automatic evaluation and human judgment show that HRAN can significantly outperform state-of-the-art models for multi-turn response generation.
	\end{abstract}
	
	\section{Introduction}
	Conversational agents include task-oriented dialog systems which are built in vertical domains for specific tasks \cite{young2013pomdp,boden2006mind,wallace2009anatomy,young2010hidden}, and non-task-oriented chatbots which aim to realize natural and human-like conversations with people regarding to a wide range of issues in open domains \cite{jafarpour2010filter}.  A common practice to build a chatbot is to learn a response generation model within an encoder-decoder framework from large scale message-response pairs \cite{shang2015neural,vinyals2015neural}.  Such models ignore conversation history when responding, which is contradictory to the nature of real conversation between humans. To resolve the problem, researchers have taken conversation history into consideration and proposed response generation for multi-turn conversation \cite{sordoni2015neural,serban2015building,serban2016multiresolution,serban2016hierarchical}. 
\begin{figure}
\includegraphics[width=6.5cm,height=5.2cm]{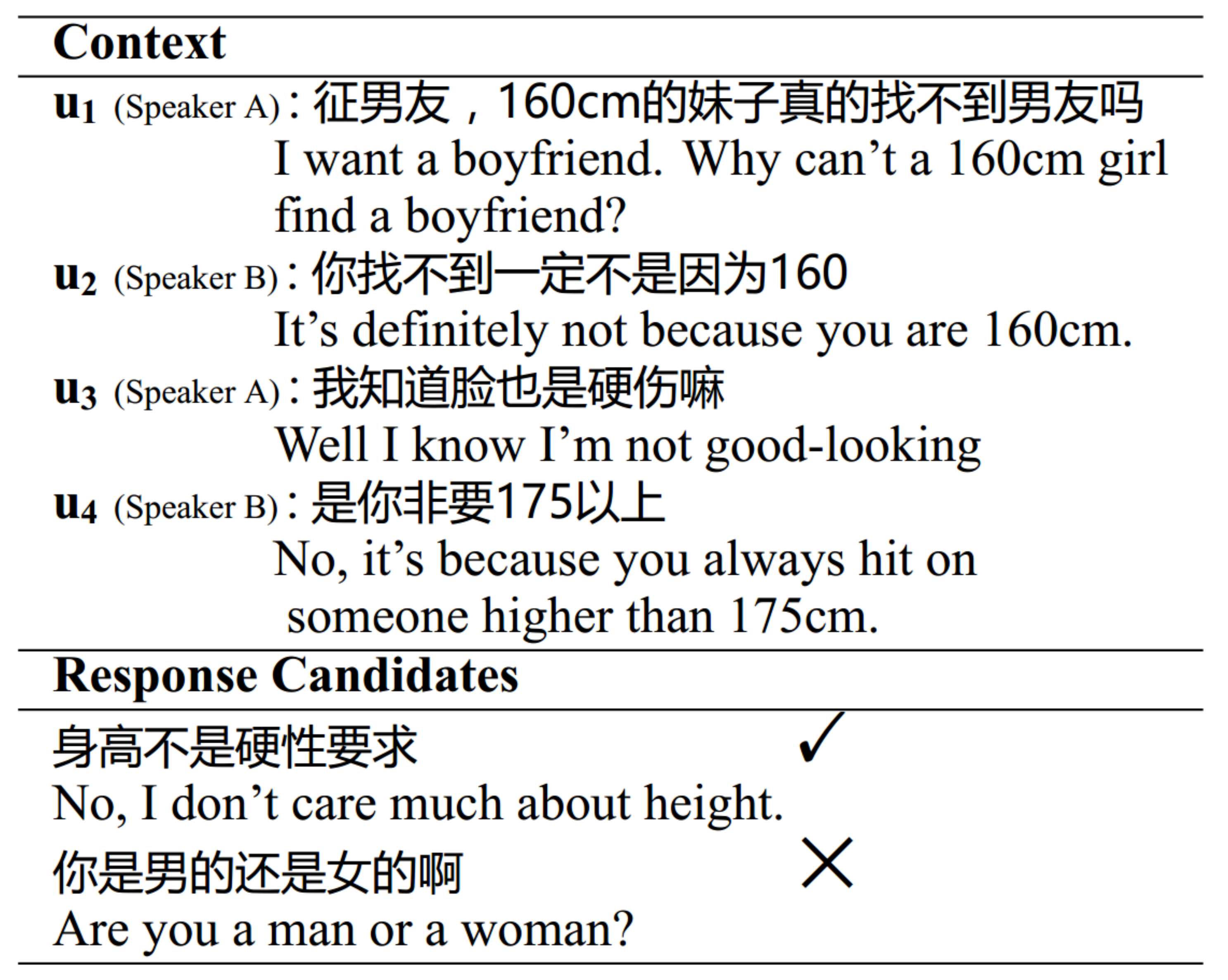}
\vspace{-5mm}
\caption{An example of multi-turn conversation}\label{fig:case_intro}
\vspace{-6mm}
\end{figure}	
	
	In this work, we study multi-turn response generation for open domain conversation in chatbots in which we try to learn a response generation model from responses and their contexts. A context refers to a message and several utterances in its previous turns.  In practice, when a message comes, the model takes the context as input and generate a response as the next turn. Multi-turn conversation requires a model to generate a response relevant to the whole context. The complexity of the task lies in two aspects: 1) a conversation context is in a hierarchical structure (words form an utterance, and utterances form the context) and has two levels of sequential relationships among both words and utterances within the structure; 2) 
	not all parts of the context are equally important to response generation. Words are differentially informative and important, and so are the utterances. 
	State-of-the-art methods such as HRED \cite{serban2016building} and VHRED \cite{serban2016hierarchical} focus on modeling the hierarchy of the context, whereas there is little exploration on how to select important parts from the context, although it is often a crucial step for generating a proper response. Without this step, existing models may lose important information in context and generate irrelevant responses\footnote{Note that one can simply concatenate all utterances and employs the classic sequence-to-sequence with attention to model word importance in generation. This method, however, loses utterance relationships and results in bad generation quality, as will be seen in expeirments.}. Figure \ref{fig:case_intro} gives an example from our data to illustrate the problem. The context is a conversation between two speakers about height and boyfriend, therefore, to respond to the context, words like ``girl'', ``boyfriend'' and numbers indicating height such as ``160'' and ``175'' are more important than ``not good-looking''. 
	Moreover, $u_1$ and $u_4$ convey main semantics of the context, and therefore are more important than the others for generating a proper response. Without modeling the word and utterance importance, the state-of-the-art model VHRED \cite{serban2016hierarchical} misses important points and gives a response ``are you a man or a woman'' which is OK if there were only $u_3$ left,  but nonsense given the  whole context. After paying attention to the important words and utterances, we can have a reasonable response like ``No, I don't care much about height'' (the response is generated by our model, as will be seen in experiments).     
 	
	We aim to model the hierarchy and the important parts of contexts in a unified framework. Inspired by the success of the attention mechanism in single-turn response generation \cite{shang2015neural}, we propose a hierarchical recurrent attention network (HRAN) for multi-turn response generation in which we introduce a hierarchical attention mechanism to dynamically highlight important parts of word sequences and the utterance sequence when generating a response. Specifically, HRAN is built in a hierarchical structure. At the bottom of HRAN, a word level recurrent neural network (RNN) encodes each utterance into a sequence of hidden vectors. In generation of each word in the response, a word level attention mechanism assigns a weight to each vector in the hidden sequence of an utterance and forms an utterance vector by a linear combination of the vectors. Important hidden vectors correspond to important parts in the utterance regarding to the generation of the word, and contribute more to the formation of the utterance vector. The utterance vectors are then fed to an utterance level RNN which constructs hidden representations of the context. Different from classic attention mechanism, the word level attention mechanism in HRAN is dependent on both the decoder and the utterance level RNN. Thus, both the current generated part of the response and the content of context can help select important parts in utterances. At the third layer, an utterance attention mechanism attends to important utterances in the utterance sequence and summarizes the sequence as a context vector. Finally, at the top of HRAN, a decoder takes the context vector as input and generates the word in the response.  HRAN mirrors the data structure in multi-turn response generation by growing from words to utterances and then from utterances to the output. It extends the architecture of current hierarchical response generation models by a hierarchical attention mechanism which not only results in better generation quality, but also provides insight into which parts in an utterance and which utterances in context contribute to response generation.  
	
	We conduct an empirical study on large scale open domain conversation data and compare our model with state-of-the-art models using both automatic evaluation and side-by-side human comparison. The results show that on both metrics our model can significantly outperform existing models for multi-turn response generation. We release our source code and data at \textit{https://github.com/LynetteXing1991/HRAN}.
	
	The contributions of the paper include (1) proposal of attending to important parts in contexts in multi-turn response generation; (2) proposal of a hierarchical recurrent attention network which models hierarchy of contexts, word importance, and utterance importance in a unified framework; (3) empirical verification of the effectiveness of the model by both automatic evaluation and human judgment.    
		\section{Related Work}	
		Most existing effort on response generation is paid to single-turn conversation. Starting from the basic sequence to sequence model \cite{sutskever2014sequence}, various models \cite{shang2015neural,vinyals2015neural,li2015diversity,xing2016topic,li2016persona, mou2016sequence} have been proposed under an encoder-decoder framework to improve generation quality from different perspectives such as relevance, diversity, and personality. Recently, multi-turn response generation has drawn attention from academia. For example, Sordoni et al. \shortcite{sordoni2015neural} proposed DCGM where context information is encoded with a multi-layer perceptron (MLP). Serban et al. \shortcite{serban2016building} proposed HRED which models contexts in a hierarchical encoder-decoder framework. Under the architecture of HRED, more variants including VHRED \cite{serban2016hierarchical} and MrRNN \cite{serban2016multiresolution} are proposed in order to introduce latent and explicit variables into the generation process. In this work, we also study multi-turn response generation. Different from the existing models which do not model word and utterance importance in generation, our hierarchical recurrent attention network simultaneously models the hierarchy of contexts and the importance of words and utterances in a unified framework.  
		
		Attention mechanism is first proposed for machine translation \cite{bahdanau2014neural,cho2015describing}, and is quickly applied to single-turn response generation afterwards \cite{shang2015neural,vinyals2015neural}. Recently, Yang et al. \shortcite{yang2016hierarchical} proposed a hierarchical attention network for document classification in which two levels of attention mechanisms are used to model the contributions of words and sentences in classification decision. Seo et al. \shortcite{seo2016hierarchical} proposed a hierarchical attention network to precisely attending objects of different scales and shapes in images. Inspired by these work, we extend the attention mechanism for single-turn response generation to a hierarchical attention mechanism for multi-turn response generation. To the best of our knowledge, we are the first who apply the hierarchical attention technique to response generation in chatbots. 	
		
		\begin{figure*}[t]
			\begin{center}
				\includegraphics[width=11.5cm,height=6.1cm]{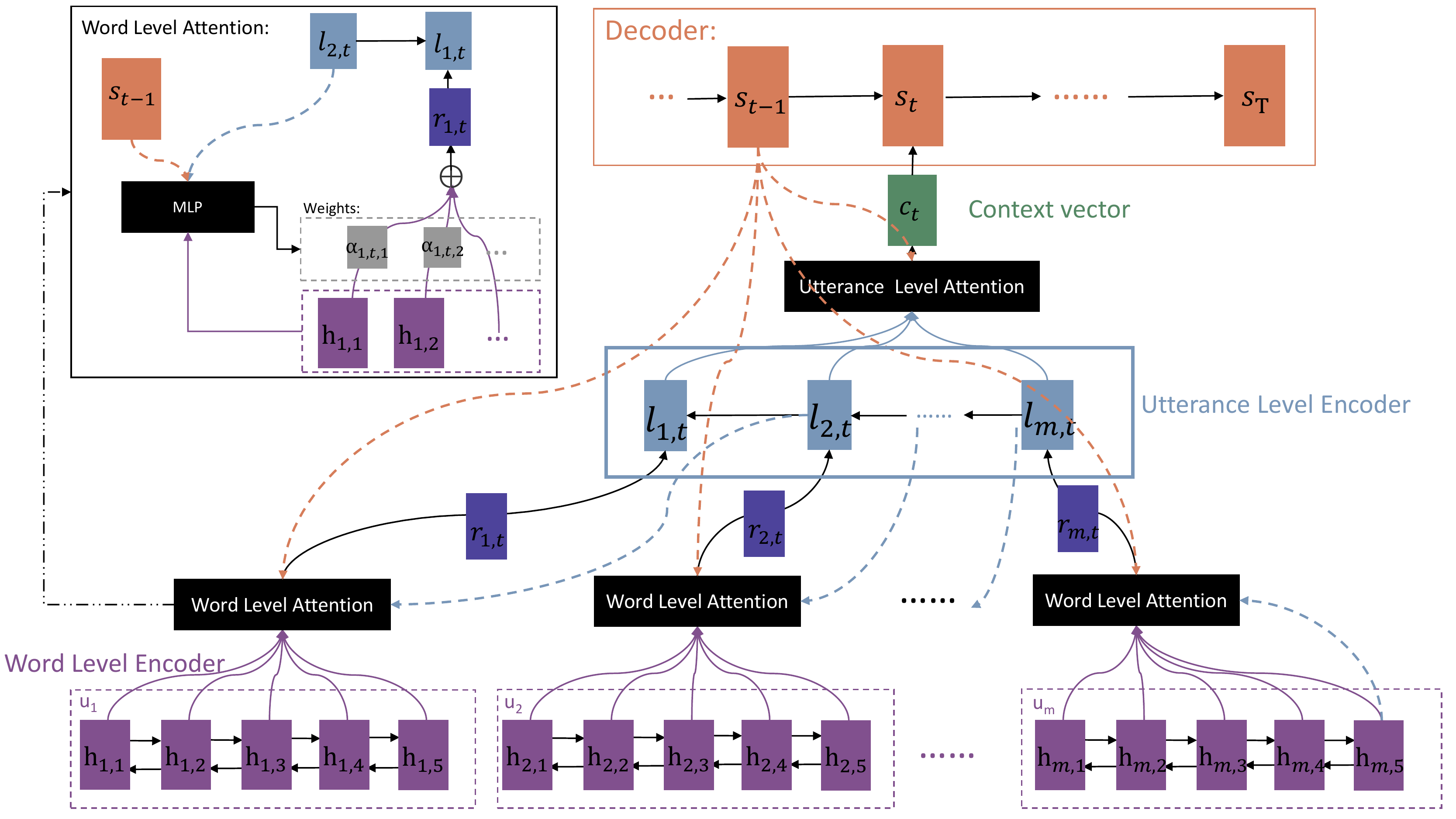}
			\end{center}
			\vspace{-4mm}		
			\caption{Hierarchical Recurrent Attention Network}\label{fig:model}
			\vspace{-4mm}		
		\end{figure*}
		
		\section{Problem Formalization}
		Suppose that we have a data set $\mathscr{D}=\{(\mathbf{U}_i, \mathbf{Y}_i)\}_{i=1}^N$. $\forall i$, $(\mathbf{U}_i,\mathbf{Y}_i)$ consists of a response $\mathbf{Y}_i=\left(y_{i,1},\ldots,y_{i,T_i}\right)$ and its context $\mathbf{U}_i=\left(u_{i,1},\ldots,u_{i,m_i}\right)$ with $y_{i,j}$ the $j$-th word, $u_{i,m_i}$ the message, and $\left(u_{i,1},\ldots,u_{i,m_i-1}\right)$ the utterances in previous turns.  In this work, we require $m_i\geqslant2$ and thus each  context has at least one utterance as conversation history. $\forall j$, $u_{i,j}=\left(w_{i,j,1}, \ldots, w_{i,j,T_{i,j}}\right)$ where $w_{i,j,k}$ is the $k$-th word.  We aim to estimate a generation probability 
		$p(y_1,\ldots,y_{T}|\mathbf{U})$ from $\mathscr{D}$, and thus given a new conversation context $\mathbf{U}$, we can generate a response $\mathbf{Y}=\left(y_1, \ldots, y_{T}\right)$ according to $p(y_1,\ldots,y_{T}|\mathbf{U})$. 
		
		In the following, we will elaborate how to construct $p(y_1,\ldots,y_{T}|\mathbf{U})$ and how to learn it.

		\section{Hierarchical Recurrent Attention Network}
		We propose a hierarchical recurrent attention network (HRAN) to model the generation probability $p(y_1,\ldots,y_{T}|\mathbf{U})$.  Figure \ref{fig:model} gives the architecture of HRAN. Roughly speaking, before generation, HRAN employs a word level encoder to encode information of every utterance in context as hidden vectors. Then, when generating every word, a hierarchical attention mechanism attends to important parts within and among utterances with word level attention and utterance level attention respectively. With the two levels of attention, HRAN works in a bottom-up way: hidden vectors of utterances are processed by the word level attention and uploaded to an utterance level encoder to form hidden vectors of the context. Hidden vectors of the context are further processed by the utterance level attention as a context vector and uploaded to the decoder to generate the word. 
		
		In the following, we will describe details and the learning objective of HRAN. 
		\subsection{Word Level Encoder}	
		Given $\mathbf{U}=\left(u_1,\ldots, u_m\right)$,  we employ a bidirectional recurrent neural network with gated recurrent units (BiGRU) \cite{bahdanau2014neural} to encode each $u_i, i\in \{1,\ldots,m\}$ as hidden vectors $\left(\mathbf{h}_{i,1},\ldots, \mathbf{h}_{i,T_i}\right)$. Formally, suppose that $u_i=\left(w_{i,1},\ldots,w_{i,T_i}\right)$, then $\forall k\in\{1,\ldots,T_i\}$, $\mathbf{h}_{i,k}$ is given by
		\begin{equation}
		\label{eq: encoder function}
		\mathbf{h}_{i,k}=concat(\overrightarrow{\mathbf{h}}_{i,k}, \overleftarrow{\mathbf{h}}_{i,k}),
		\end{equation}
		where $concat(\cdot,\cdot)$ is an operation defined as concatenating the two arguments together, $\overrightarrow{\mathbf{h}}_{i,k}$ is the $k$-th hidden state of a forward GRU \cite{cho2014properties}, and $\overleftarrow{\mathbf{h}}_{i,k}$ is the $k$-th hidden state of a backward GRU. The forward GRU reads $u_i$ in its order (i.e., from $w_{i,1}$ to $w_{i,T_i}$), and calculates $\overrightarrow{\mathbf{h}}_{i,k}$ as 
		\begin{equation}
		\label{GRU unit}
		\small
		\vspace{-0.7mm}
		\begin{aligned}
		&\mathbf{z}_k=\sigma(\mathbf{W}_z\mathbf{e}_{i,k}+\mathbf{V}_z\overrightarrow{\mathbf{h}}_{i,k-1})\\
		&\mathbf{r}_k=\sigma(\mathbf{W}_r\mathbf{e}_{i,k}+\mathbf{V}_r\overrightarrow{\mathbf{h}}_{i,k-1})\\
		&\mathbf{s}_k=tanh(\mathbf{W}_s\mathbf{e}_{i,k}+\mathbf{V}_s(\overrightarrow{\mathbf{h}}_{i,k-1} \circ \mathbf{r}_k))\\
		&\overrightarrow{\mathbf{h}}_{i,k}=(1-\mathbf{z}_k)\circ \mathbf{s}_k+\mathbf{z}_k\circ\overrightarrow{\mathbf{h}}_{i,k-1},
		\end{aligned}
		\vspace{-0.7mm}
		\end{equation}
		where $\overrightarrow{\mathbf{h}}_{i,0}$ is initialized with a isotropic Gaussian distribution, $\mathbf{e}_{i,k}$ is the embedding of $w_{i,k}$, $\mathbf{z}_k$ and $\mathbf{r}_k$ are an update gate and a reset gate respectively, $\sigma(\cdot)$ is a sigmoid function, and $\mathbf{W}_z, \mathbf{W}_r, \mathbf{W}_s, \mathbf{V}_z, \mathbf{V}_r, \mathbf{V}_s$ are parameters.  The backward GRU reads $u_i$ in its reverse order (i.e., from $w_{i,T_i}$ to $w_{i,1}$) and generates $\{\overleftarrow{\mathbf{h}}_{i,k}\}_{k=1}^{T_i}$ with a parameterization similar to the forward GRU.

		\subsection{Hierarchical Attention and Utterance Encoder}
		Suppose that the decoder has generated $t-1$ words, at step $t$, word level attention calculates a weight vector $\left(\alpha_{i,t,1}, \ldots, \alpha_{i,t,T_i}\right)$ (details are described later) for $\{\mathbf{h}_{i,j}\}_{j=1}^{T_i}$ and represents utterance $u_i$ as 
		a vector $\mathbf{r}_{i,t}$. $\forall i\in \{1,\ldots, m\}$, $\mathbf{r}_{i,t}$ is defined by
		\begin{equation}
		\vspace{-1mm}
		\small
		\label{wordsum}
		\mathbf{r}_{i,t}=\sum_{j=1}^{T_i}\alpha_{i,t,j}\mathbf{h}_{i,j}.
		\vspace{-0.3mm}
		\end{equation}
		$\{\mathbf{r}_{i,t}\}_{i=1}^m$ are then utilized as input of an utterance level encoder and transformed to $\left(\mathbf{l}_{1,t}, \ldots, \mathbf{l}_{m,t}\right)$ as hidden vectors of the context. After that, utterance level attention assigns a weight $\beta_{i,t}$ to $ \mathbf{l}_{i,t}$ (details are described later) and forms a context vector $\mathbf{c}_t$ as
		\begin{equation}
		\small
		\label{utterancesum}
		\mathbf{c}_t=\sum_{i=1}^{m}\beta_{i,t}\mathbf{l}_{i,t}.
		\end{equation}
		In both Equation (\ref{wordsum}) and Equation (\ref{utterancesum}), the more important a hidden vector is, the larger weight it will have, and the more contributions it will make to the high level vector (i.e., the utterance vector and the context vector). This is how the two levels of attention attends to the important parts of utterances and the important utterances in generation. 
		
		More specifically, the utterance level encoder is a backward GRU which processes $\{\mathbf{r}_{i,t}\}_{i=1}^m$ from the message $\mathbf{r}_{m,t}$ to the earliest history $\mathbf{r}_{1,t}$. Similar to Equation (\ref{GRU unit}), $\forall i\in \{m,\dots, 1\}$, $\mathbf{l}_{i,t}$ is calculated as 
		\begin{equation}
		\label{BackGRU}
		\small
		\begin{aligned}
		&\mathbf{z}'_i=\sigma(\mathbf{W}_{zl}\mathbf{r}_{i,t}+\mathbf{V}_{zl} \mathbf{l}_{i+1,t})\\
		&\mathbf{r}'_i=\sigma(\mathbf{W}_{rl}\mathbf{r}_{i,t}+\mathbf{V}_{rl} \mathbf{l}_{i+1,t})\\
		&\mathbf{s}'_i=tanh(\mathbf{W}_{sl}\mathbf{r}_{i,t}+\mathbf{V}_{sl}(\mathbf{l}_{i+1,t} \circ \mathbf{r}'_i))\\
		&\mathbf{l}_{i,t}=(1-\mathbf{z}'_i)\circ \mathbf{s}'_i+\mathbf{z}'_i\circ \mathbf{l}_{i+1,t},
		\end{aligned}
		\end{equation}
		where $\mathbf{l}_{m+1,t}$ is initialized with a isotropic Gaussian distribution, $\mathbf{z}'_i$ and $\mathbf{r}'_i$ are the update gate and the reset gate of the utterance level GRU respectively, and $\mathbf{W}_{zl}, \mathbf{V}_{zl}, \mathbf{W}_{rl}, \mathbf{V}_{rl}, \mathbf{W}_{sl}, \mathbf{V}_{sl}$ are parameters. 
		
		Different from the classic attention mechanism, word level attention in HRAN depends on both the hidden states of the decoder and the hidden states of the utterance level encoder. It works in a reverse order by first weighting $\{\mathbf{h}_{m,j}\}_{j=1}^{T_m}$ and then moving towards $\{\mathbf{h}_{1,j}\}_{j=1}^{T_1}$ along the utterance sequence. $\forall i\in \{m,\ldots,1\}, j\in \{1, \ldots, T_i\}$, weight $\alpha_{i,t,j}$ is calculated as
		\begin{equation}
		\label{wordatt}
		\small
		\begin{aligned}
		&e_{i,t,j}= \eta(\mathbf{s}_{t-1},\mathbf{l}_{i+1,t},\mathbf{h}_{i,j});\\
		&\alpha_{i,t,j}=\frac{exp(e_{i,t,j})}{\sum_{k=1}^{T_i} exp(e_{i,t,k})},
		\end{aligned}
		\end{equation}
		where $\mathbf{l}_{m+1,t}$ is initialized with a isotropic Gaussian distribution, $\mathbf{s}_{t-1}$ is the $(t-1)$-th hidden state of the decoder, and $\eta(\cdot)$ is a multi-layer perceptron (MLP) with tanh as an activation function. 
		
		Note that the word level attention and the utterance level encoding are dependent with each other and alternatively conducted (first attention then encoding). The motivation we establish the dependency between $\alpha_{i,t,j}$ and $\mathbf{l}_{i+1,t}$ is that content from the context (i.e., $\mathbf{l}_{i+1,t}$) could help identify important information in utterances, especially when $\mathbf{s}_{t-1}$ is not informative enough (e.g., the generated part of the response are almost function words). We require the utterance encoder and the word level attention to work reversely, because we think that compared to history, conversation that happened after an utterance in the context is more likely to be capable of identifying important information in the utterance for generating a proper response to the context. 
		
		With $\{\mathbf{l}_{i,t}\}_{i=1}^m$, the utterance level attention calculates a weight $\beta_{i,t}$ for $ \mathbf{l}_{i,t}$ as 
		\begin{equation}
		\label{UttAtt}
		\small
		\begin{aligned}
		&e'_{i,t}= \eta(\mathbf{s}_{t-1},\mathbf{l}_{i,t});\\
		&\beta_{i,t}=\frac{exp(e'_{i,t})}{\sum_{i=1}^{m} exp(e'_{i,t})}.\\
		\end{aligned}
		\end{equation}
		
		\subsection{Decoding the Response}
		The decoder of HRAN is a RNN language model \cite{mikolov2010recurrent} conditioned on the context vectors $\{\mathbf{c}_t\}_{t=1}^{T}$ given by Equation (\ref{utterancesum}). Formally, the probability distribution $p(y_1,\ldots,y_{T}|\mathbf{U})$ is defined as
		\begin{equation}
		\vspace{-1mm}
		\small
		p(y_1,...,y_{T}|\mathbf{U})=p(y_1|\mathbf{c}_1)\prod_{t=2}^{T}p(y_t|\mathbf{c}_t,y_1,...,y_{t-1}).
		\vspace{-0.5mm}
		\end{equation}
		where $p(y_t|\mathbf{c}_t,y_1,...,y_{t-1})$ is given by
		\begin{equation}
		\label{eq: decoder function}
		\small
		\begin{aligned}
		&\mathbf{s}_t=f(\mathbf{e}_{y_{t-1}},\mathbf{s}_{t-1},\mathbf{c}_t)\\
		&p(y_t|\mathbf{c}_t,y_1,...,y_{t-1})=\mathbbm{I}_{y_t} \cdot softmax(\mathbf{s}_t,\mathbf{e}_{y_{t-1}}),
		\end{aligned}
		\end{equation}
		where $\mathbf{s}_t$ is the hidden state of the decoder at step $t$, $\mathbf{e}_{y_{t-1}}$ is the embedding of $y_{t-1}$, $f$ is a GRU, $\mathbbm{I}_{y_t}$ is the one-hot vector for $y_t$, and $softmax(\mathbf{s}_t,\mathbf{e}_{y_{t-1}})$ is a $V$-dimensional vector with $V$ the response vocabulary size and each element the generation probability of a word. In practice, we employ the beam search \cite{tillmann2003word} technique to generate the $n$-best responses.  
		
		Let us denote $\Theta$ as the parameter set of HRAN, then we estimate $\Theta$ from $\mathscr{D}=\{(\mathbf{U}_i, \mathbf{Y}_i)\}_{i=1}^N$ by minimizing the following objective function:
		\begin{equation}
		\small
		\hat{\Theta}=\underset{\Theta}{\arg\min} \thickspace -\sum_{i=1}^N \log\left(p(y_{i,1},...,y_{i,T_i}|\mathbf{U_i})\right)
		\end{equation}
		
		\section{Experiments}
		We compared HRAN with state-of-the-art methods by both automatic evaluation and side-by-side human judgment.
		\subsection{Data Set}
		We built a data set from Douban Group\footnote{https://www.douban.com/group/explore} which is a popular Chinese social networking service (SNS) allowing users to discuss a wide range of topics in groups through posting and commenting. In Douban Group, regarding to a post under a specific topic, two persons can converse with each other by one posting a comment and the other quoting it and posting another comment. We crawled $20$ million conversations between two persons with the average number of turns as  $6.32$. The data covers many different topics and can be viewed as a simulation of open domain conversations in a chatbot. In each conversation, we treated the last turn as response, and the remaining turns as context. As preprocessing, we first employed Stanford Chinese word segmenter\footnote{http://nlp.stanford.edu/software/segmenter.shtml} to tokenize each utterance in the data. Then we removed the conversations whose response appearing more than $50$ times in the whole data to prevent them from dominating learning. We also removed the conversations shorter than $3$ turns and the conversations with an utterance longer than $50$ words. After the preprocessing, there are $1,656,652$ conversations left. From them, we randomly sampled $1$ million conversations as training data, $10,000$ conversations as validation data, and $1,000$ conversations as test data, and made sure that there is no overlap among them.  In the test data, the contexts were used to generate responses and their responses were used as ground truth to calculate perplexity of generation models. We kept the $40,000$ most frequent words in the contexts of the training data to construct a context vocabulary. The vocabulary covers $98.8\%$ of words appearing in the contexts of the training data. Similarly, we constructed a response vocabulary that contains the $40,000$ most frequent words in the responses of the training data which covers $99.0\%$ words appearing in the responses. Words outside the two vocabularies were treated as ``UNK''. The data will be publicly available.
		
		\subsection{Baselines}
		We compared HRAN with the following models:
		
		\textbf{S2SA}: we concatenated all utterances in a context as a long sequence and treated the sequence and the response as a message-response pair. By this means, we transformed the problem of multi-turn response generation to a problem of single-turn response generation and employed the standard sequence to sequence with attention \cite{shang2015neural} as a baseline. 
		
		\textbf{HRED}: the hierarchical encoder-decoder model proposed by \cite{serban2016building}.
		
		\textbf{VHRED}: a modification of HRED \cite{serban2016hierarchical} where latent variables are introduced in to generation.
\begin{figure*}[t]
	\includegraphics[width=16cm,height=8.3cm]{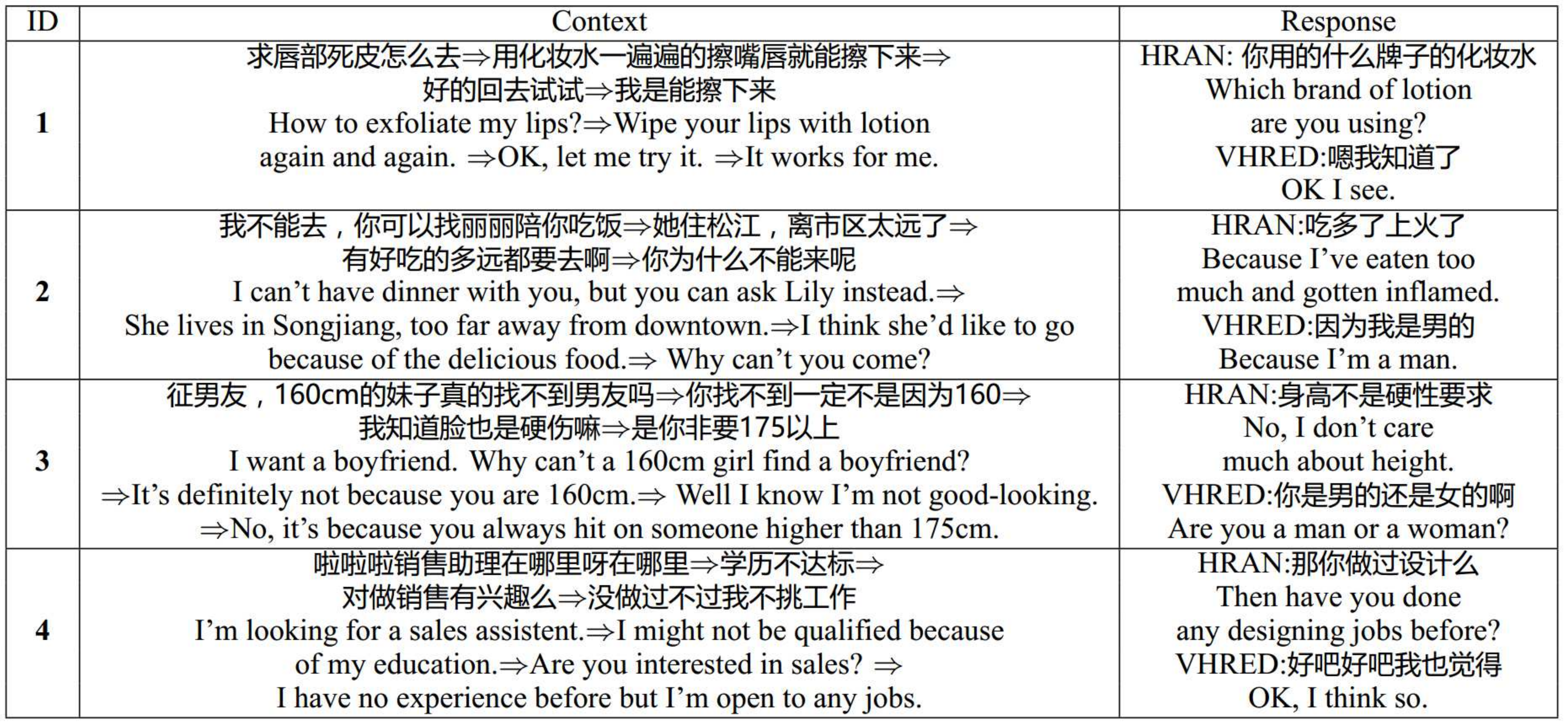}
	\vspace{-7mm}
	\caption{Case study (utterances between two persons in contexts are split by ``$\Rightarrow$'')}\label{fig:case_study}
	\vspace{-5.5mm}
\end{figure*}	
		In all models, we set the dimensionality of hidden states of encoders and decoders as $1000$, and the dimensionality of word embedding as $620$. All models were initialized with isotropic Gaussian distributions $\mathcal{X} \sim \mathcal{N}(0,0.01)$ and trained with an AdaDelta algorithm \cite{zeiler2012adadelta} on a NVIDIA Tesla K40 GPU. The batch size is $128$. We set the initial learning rate as $1.0$ and reduced it by half if the perplexity on validation began to increase. We implemented the models with an open source deep learning tool Blocks\footnote{\url{https://github.com/mila-udem/blocks}}. 
		
		\subsection{Evaluation Metrics}
		How to evaluate a response generation model is still an open problem but not the focus of the paper. We followed the existing work and employed the following metrics:
		\begin{table}
			\small
			\centering
			\begin{tabular}{|c|c|c|}
				\hline
				Model &Validation Perplexity &Test Perplexity\\
				\hline
				S2SA &43.679&44.508\\
				\hline
				HRED &46.279&47.467\\
				\hline
				VHRED &44.548&45.484\\
				\hline
				HRAN &40.257&\textbf{41.138}\\
				\hline		
			\end{tabular}
			\vspace{-2mm}		
			\caption{Perplexity results}
			\label{tab:ppl}
			\vspace{-6.0mm}		
		\end{table}
		
		\textbf{Perplexity}: following \cite{vinyals2015neural}, we employed perplexity as an evaluation metric.  Perplexity is defined by Equation (\ref{perplexity}). It measures how well a model predicts human responses. Lower perplexity generally indicates better generation performance. In our experiments, perplexity on validation was used to determine when to stop training.  If the perplexity stops decreasing and the difference is smaller than $2.0$ five times in validation, we think that the algorithm has reached convergence and terminate training. We tested the generation ability of different models by perplexity on the test data.
		\begin{equation}
		\label{perplexity}
		\vspace{-3mm}
		\small
		PPL=exp\left\{-\frac{1}{N}\Sigma_{i=1}^{N}\log(p(\mathbf{Y}_i|\mathbf{U}_i))\right\}.
		\vspace{-2mm}
		\end{equation}
		\begin{table}
			\small
			\centering
			\begin{tabular}{|c|c|c|c|c|}
				\hline
				Models &Win &Loss &Tie &Kappa\\\hline
				HRAN v.s. S2SA &\textbf{27.3} &20.6 &52.1& 0.37\\
				\hline
				HRAN v.s. HRED &\textbf{27.2} &21.2 &51.6& 0.35\\
				\hline
				HRAN v.s. VHRED &\textbf{25.2} &20.4 &54.4& 0.34\\
				\hline		
			\end{tabular}
			\vspace{-2mm}
			\caption{Human annotation results (in \%)}
			\label{tab:human}
			\vspace{-5mm}
		\end{table}

		\textbf{Side-by-side human annotation}: we also compared HRAN with every baseline model by side-by-side human comparison. Specifically, we recruited three native speakers with rich Douban Group experience as human annotators. To each annotator, we showed a context of a test example with two generated responses, one from HRAN and the other one from a baseline model. Both responses are the top one results in beam search. The two responses were presented in random order. We then asked the annotator to judge which one is better. The criteria is, response A is better than response B if (1) A is relevant, logically consistent to the context, and fluent, while B is either irrelevant or logically contradictory to the context, or it is disfluent (e.g., with grammatical errors or UNKs); or (2) both A and B are relevant, consistent, and fluent, but A is more informative and interesting than B (e.g., B is a universal reply like ``I see''). If the annotator cannot tell which one is better, he/she was asked to label a ``tie''. Each annotator individually judged $1000$ test examples for each HRAN-baseline pair, and in total, each one judged $3000$ examples (for three pairs). Agreements among the annotators were calculated using Fleiss' kappa \cite{fleiss1973equivalence}.

		Note that we do not choose BLEU \cite{papineni2002bleu} as an evaluation metric, because (1) Liu et al. \cite{liu2016not} have proven that BLEU is not a proper metric for evaluating conversation models as there is weak correlation between BLEU and human judgment; (2) different from the single-turn case, in multi-turn conversation, one context usually has one copy in the whole data. Thus, without any human effort like what Sordoni et al. \cite{sordoni2015neural} did in their work, each context only has a single reference in test. This makes BLEU even unreliable as a measurement of generation quality in open domain conversation due to the diversity of responses.  
		
		\subsection{Evaluation Results}
		
		Table \ref{tab:ppl} gives the results on perplexity. HRAN achieves the lowest perplexity on both validation and test. We conducted t-test on test perplexity and the result shows that the improvement of HRAN over all baseline models is statistically significant (p-value $< 0.01$).

		Table \ref{tab:human} shows the human annotation results. The ratios were calculated by combining the annotations from the three judges together. We can see that HRAN outperforms all baseline models and all comparisons have relatively high kappa scores which indicates that the annotators reached relatively high agreements in judgment. Compared with S2SA, HRED, and VHRED, HRAN achieves preference gains (win-loss) $6.7$\%, $6$\%, $4.8$\% respectively. Sign test results show that the improvement is statistically significant (p-value $< 0.01$ for HRAN v.s. S2SA and HRAN v.s. HRED, and p-value $<0.05$ for HRAN v.s. VHRED).  Among the three baseline models, S2SA is the worst one, because it loses relationships among utterances in the context.  VHRED is the best baseline model,  which is consistent with the existing literatures \cite{serban2016hierarchical}.  We checked the cases on which VHRED loses to HRAN and found that on $56$\% cases, VHRED generated irrelevant responses while responses given by HRAN are relevant, logically consistent, and fluent.

		\subsection{Discussions}
		\textbf{Case study:} Figure \ref{fig:case_study} lists some cases from the test set to compare HRAN with the best baseline VHRED. We can see that HRAN  not only can answer the last turn in the context (i.e., the message) properly by understanding the context  (e.g., case 2), but also be capable of starting a new topic according to the conversation history to keep the conversation going (e.g., case 1). In case 2, HRAN understands that the message is actually asking ``why can't you come to have dinner with me?'' and generates a proper response that gives a plausible reason. In case 1, HRAN properly brings up a new topic by asking the ``brand'' of the user's ``lotion'' when the current topic ``how to exfoliate my skin'' has come to an end. The new topic is based on the content of the context and thus can naturally extends the conversation in the case.
		
		\begin{figure*}[!t]\vspace{-4mm}
			\centering
			\subfigure[Visualization of case 1]{
				\label{fig:4860} 
				\includegraphics[width=5.5cm,height=3.7cm]{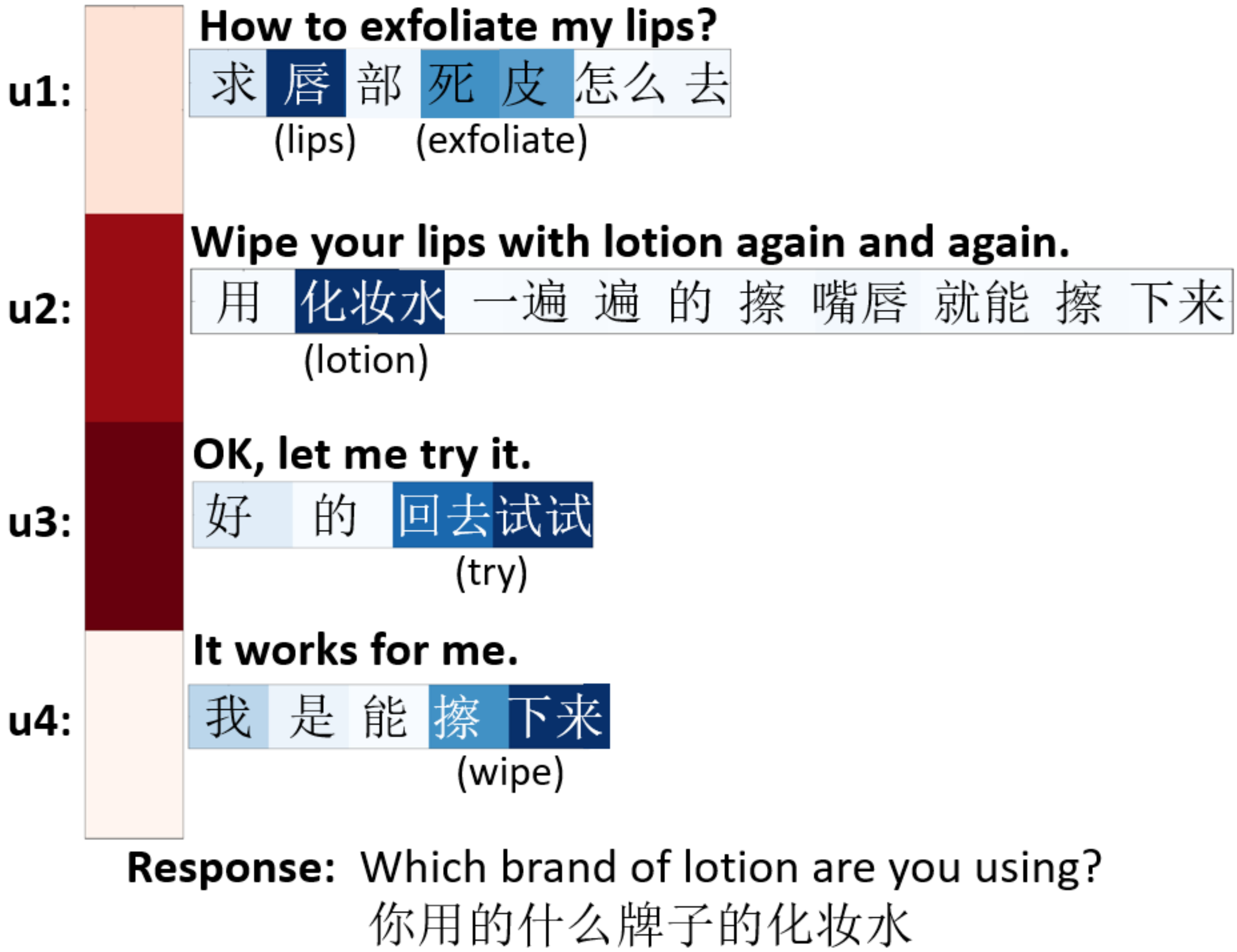}}
			\subfigure[Visualization of case 2]{
				\label{fig:4535} 
				\includegraphics[width=5.7cm,height=3.7cm]{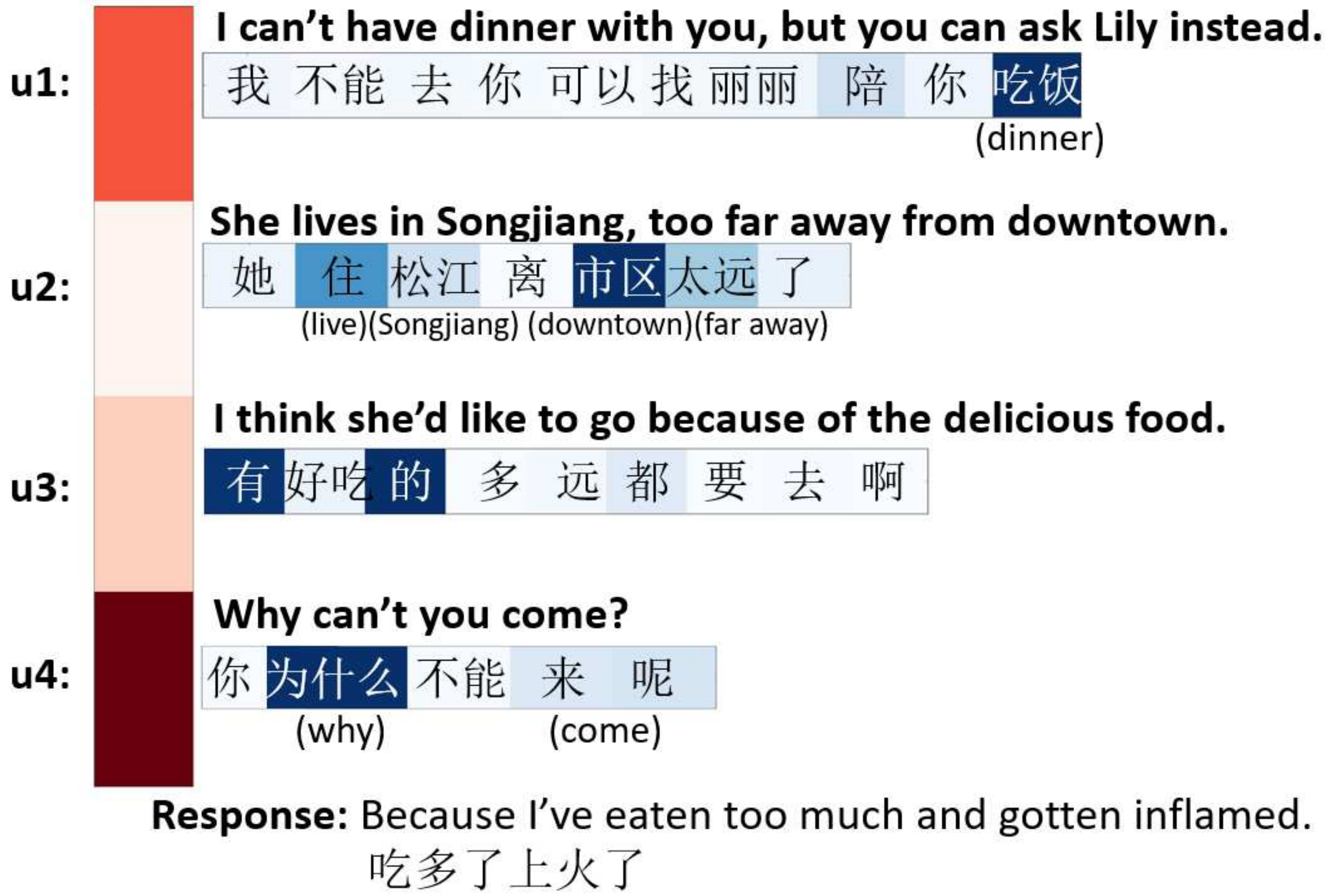}}
			\subfigure[Visualization of case 3]{
				\label{fig:4620}
				\includegraphics[width=6.0cm,height=3.7cm]{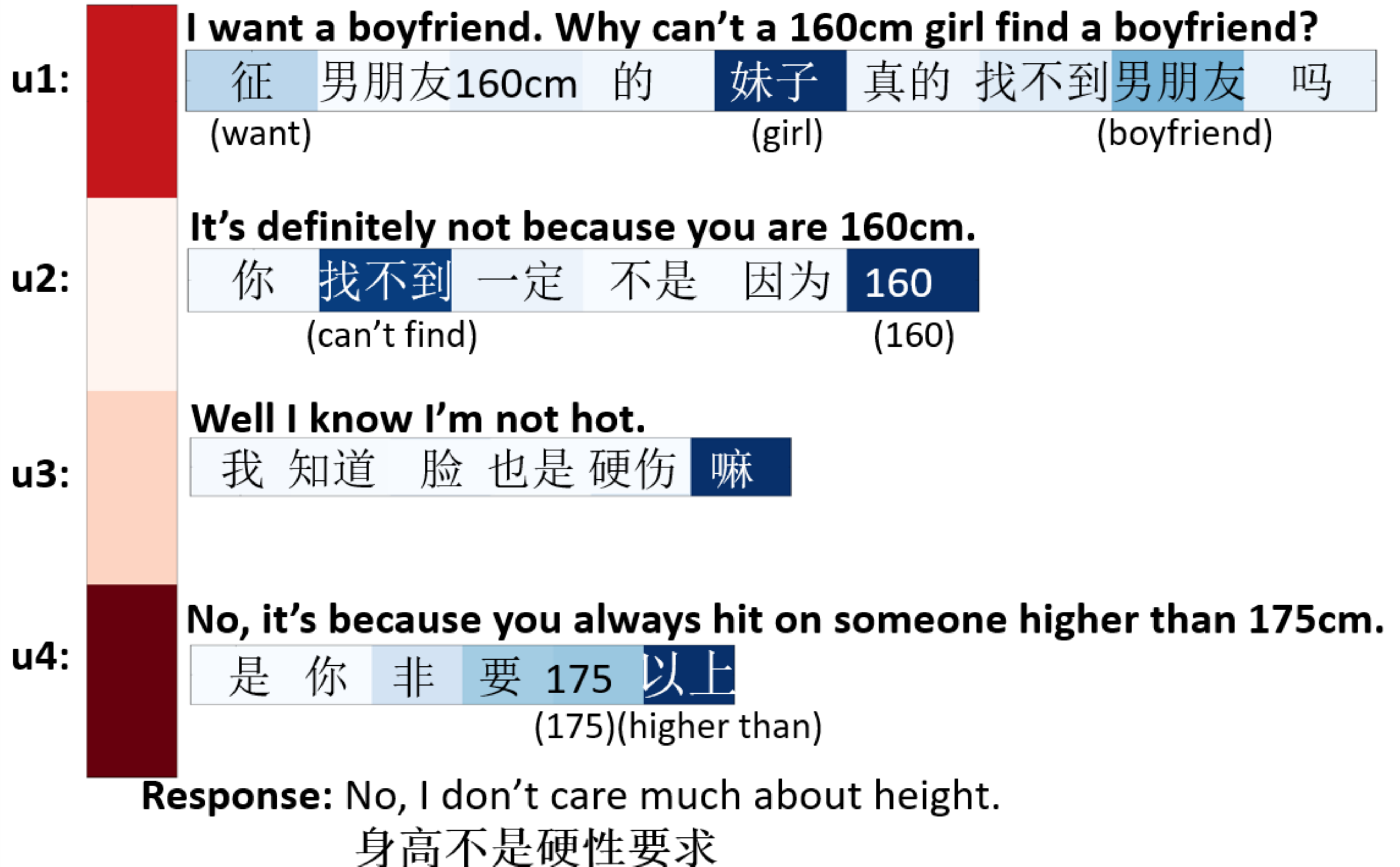}}
			\subfigure[Visualization of case 4]{
				\label{fig:4408}
				\includegraphics[width=5.8cm,height=3.7cm]{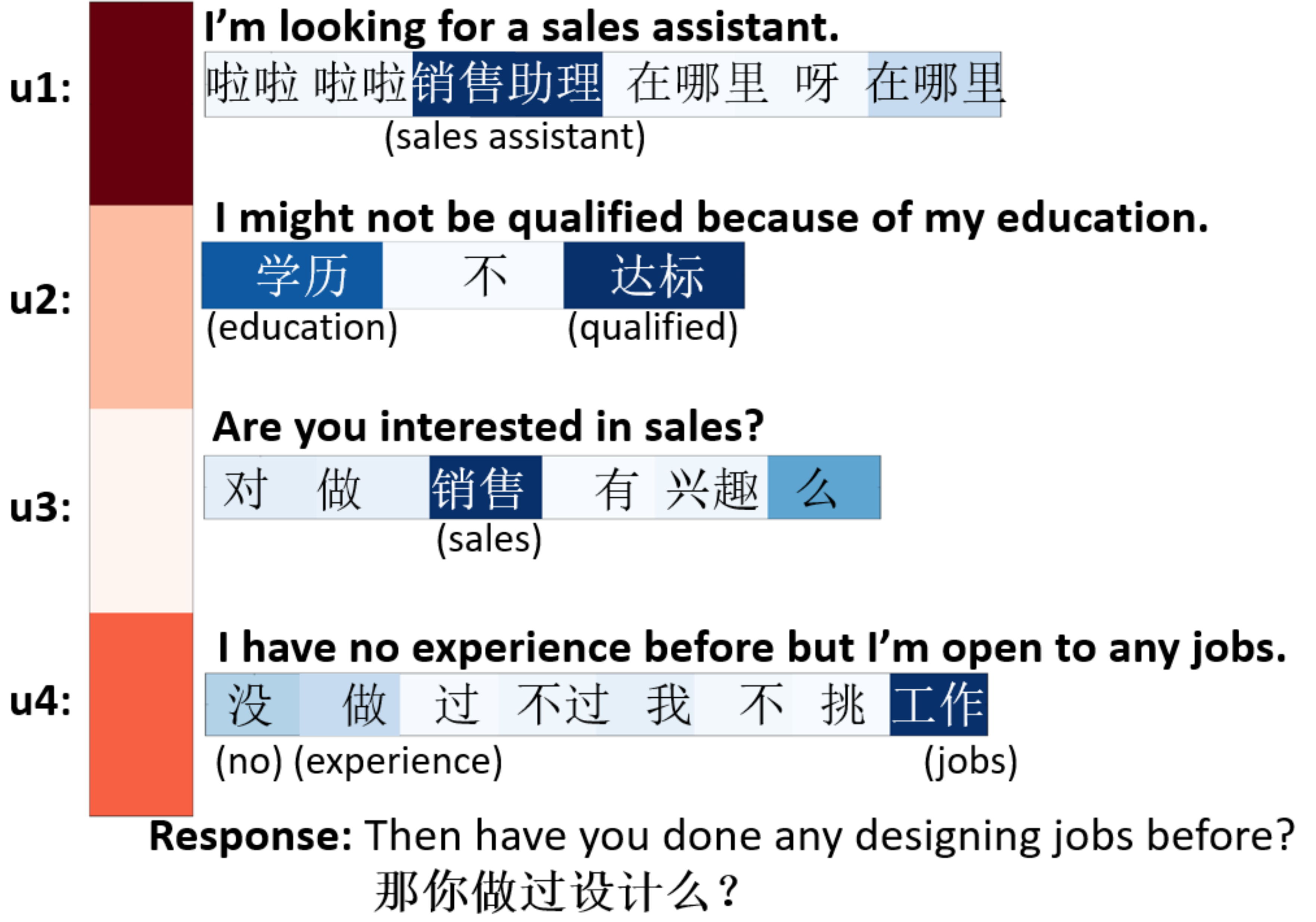}}
			\vspace{-4mm}
			\caption{Attention visualization (the importance of a word or an utterance is calculated as their average weights when generating the whole response)}
			\label{fig:visualization} 
			\vspace{-4mm}
		\end{figure*} 
		%
		%
		
		\begin{table}
			\small
			\centering
			\begin{tabular}{|c|c|c|c|c|}
				\hline
				Model &Win &Loss &Tie&PPL\\\hline
				No UD Att&22.3\% &24.8\% &52.9\%&41.54\\
				\hline
				No Word Att&20.4\%&25.0\%&50.6\%&43.24\\
				\hline
				No Utterance Att& 21.1\%&23.7\%&55.2\%&47.35\\
				\hline		
			\end{tabular}
			\vspace{-2mm}
			\caption{Model ablation results}
			\label{tab:model ablation}
			\vspace{-5.5mm}
		\end{table}
		\textbf{Visualization of attention:} to further illustrate why HRAN can generate high quality responses, we visualized the hierarchical attention for the cases in Figure \ref{fig:case_study} in Figure \ref{fig:visualization}. In every sub-figure, each line is an utterance with blue color indicating word importance. The leftmost column of each sub-figure uses red color to indicate utterance importance. Darker color means more important words or utterances. The importance of a word or an utterance was calculated by the average weight of the word or the utterance assigned by attention in generating the response given at the bottom of each sub-figure. It reflects an overall contribution of the word or the utterance to generate the response.  Above each line, we gave the translation of the utterance, and below it, we translated important words. Note that word-to-word translation may cause confusion sometimes, therefore, we left some words (most of them are function words) untranslated. We can see that the hierarchical attention mechanism in HRAN can attend to both important words and important utterances in contexts. For example, in Figure \ref{fig:4620}, words including ``girl'' and ``boyfriend'' and numbers including ``160'' and ``175'' are highlighted, and $u_1$ and $u_4$ are more important than others. The result matches our intuition in introduction.  In Figure \ref{fig:4535}, HRAN assigned larger weights to $u_1$, $u_4$ and words like ``dinner'' and ``why''.  This explains why the model can understand that the message is actually asking ``why can't you come to have dinner with me?''. The figures provide us insights on how HRAN understands contexts in generation.
		
		\textbf{Model ablation:} we then examine the effect of different 
		components of HRAN by removing them one by one. We first removed $\mathbf{l}_{i+1}$ from  $\eta(\mathbf{s}_{t-1},\mathbf{l}_{i+1,t},\mathbf{h}_{i,j})$ in Equation (\ref{wordatt}) (i.e., removing utterance dependency from word level attention) and denoted the model as ``No UD Att'', then we removed word level attention and utterance level attention separately, and denoted the models as ``No Word Att'' and ``No Utterance Att'' respectively. We conducted side-by-side human comparison on these models with the full HRAN  on the test data and also calculated their test perplexity (PPL).  Table \ref{tab:model ablation} gives the results. We can see that all the components are useful because removing any of them will cause performance drop. Among them, word level attention is the most important one as HRAN achieved the most preference gain ($4.6$\%) to No Word Att on human comparison.

		\textbf{Error analysis:} we finally investigate how to improve HRAN in the future by analyzing the cases on which HRAN loses to VHRED. The errors can be summarized as: $51.81$\% logic contradiction, $26.95$\% universal reply, $7.77$\% irrelevant response, and $13.47$\% others. Most bad cases come from universal replies and responses that are logically contradictory to contexts. This is easy to understand as HRAN does not explicitly model the two issues. The result also indicates that (1) although contexts provide more information than single messages, multi-turn response generation still has the ``safe response'' problem as the single-turn case; (2) although attending to important words and utterances in generation can lead to informative and logically consistent responses for many cases like those in Figure \ref{fig:case_study}, it is still not enough for fully understanding contexts due to the complex nature of conversations. The irrelevant responses might be caused by wrong attention in generation. Although the analysis might not cover all bad cases (e.g., HRAN and VHRED may both give bad responses), it sheds light on our future directions: (1) improving response diversity, e.g., by introducing extra content into generation like Xing et al. \cite{xing2016topic} and Mou et al. \cite{mou2016sequence} did for single-turn conversation; (2) modeling logics in contexts; (3) improving attention. 
		
		\section{Conclusion}
		We propose a hierarchical recurrent attention network (HRAN) for multi-turn response generation in chatbots. Empirical studies on large scale conversation data show that HRAN can significantly outperform state-of-the-art models.  
		
\bibliography{acl2016}
\bibliographystyle{acl2016}

\appendix

\end{document}